\documentclass[11pt]{article}
\usepackage[final]{acl}
\usepackage{times}
\usepackage{latexsym}
\usepackage[T1]{fontenc}
\usepackage[utf8]{inputenc}
\usepackage{amsmath}
\usepackage{microtype}
\usepackage{booktabs}
\usepackage{graphicx}

\title{You Can Tell Who's Asking: What the Web's Questions\\Are Made Of, and Where They Come From}
\author{%
  Calvin Zhou\textsuperscript{1,2}\thanks{Work done during an internship at Bodhium Labs.} \quad
  Vincent McCloskey\textsuperscript{2} \quad
  Krishna Srinivasan\textsuperscript{2} \\[2pt]
  \textsuperscript{1}University of Pennsylvania \quad
  \textsuperscript{2}Bodhium Labs \\[2pt]
  \texttt{\{calvin,\,vincent,\,krishna\}@bodhiumlabs.com}%
}

\begin{document}
\maketitle

\begin{abstract}
Questions scraped from the web are used across academia and industry as a proxy for what people want to know. Across QA training data, retrieval benchmarks, and content strategy, questions on a page are assumed to reflect human intent. We test this assumption at scale by extracting 13.4B question occurrences across 110 FineWeb snapshots (2013-2025), and report three findings. First, you can tell who is asking: provenance (the host/page of questions) leaves a signal in question form, and a logistic model can separate genuine user questions from templated/manufactured ones at AUC 0.725 via length and surrounding context rather than question type, though only 0.554 against commerce FAQ writing. Second, question frequency does not measure demand: the most-frequent questions are boilerplate/templated (over 70\% of the top thousand), so occurrence counts measure how often a string was published and not how often it was asked. Third, over twelve years the genuine share of occurrences fell by 79\% (42-56\% after controlling for crawl composition), with question length and context decreasing. We present the first diachronic, occurrence-level measurement of web question provenance, and find the crawlable web's questions have shifted from being asked by humans toward manufactured for machines to read.
\end{abstract}

\section{Introduction}

A question mark on a page was never a guarantee that someone genuinely wanted an answer, but it was a reasonable bet. For years, that assumption has weakened as search engines rewarded pages for including questions. FAQ blocks were attached to the bottom of pages to rank higher, with a rise evident in \emph{FAQPage} markup adoption \citep{brinkmann2023wdc,webalmanac2024structureddata}. That reward has diminished since then \citep{webalmanac2024structureddata}, but in the Generative Engine Optimization (GEO) era, content more than ever is being written for machines \citep{aggarwal2024geo}, with machine-generated text making up a growing share of the crawlable web since late 2022 \citep{he2026degentweb,dolezal2026aitext}. Yet, extracted web questions are still seen as what people want to know in QA training data \citep{huber2022ccqa,dinzinger2025webfaq}, retrieval benchmarks \citep{nguyen2016msmarco,kwiatkowski2019nq}, and content strategy.

What a question dataset can represent is dependent on its source \citep{gebru2021datasheets,dodge2021c4}. Datasets of genuine questions, such as query logs and community-QA archives, are small and kept in-house, with the AOL log with true query frequencies even being recalled \citep{pass2006aol}. Web-scale datasets, whether FAQ-markup mined or autocomplete-harvested, include both the web's genuine and manufactured sources (App.~\ref{app:landscape}). Mining full page text at least provides occurrence counts, but a count of appearances is not a count of asks. The scarce but valuable signal is \emph{popularity}, which turns a list of questions into a distribution.

We measure the web's question layer from 2013-2025 across FineWeb \citep{penedo2024fineweb}, with \emph{question occurrence} as our unit: a single question-bearing sentence on a crawled page. We ask whether a question's provenance is determinable from its form (\S\ref{sec:interaction}), whether how often it appears relates to demand (\S\ref{sec:freqdemand}), and how the proportions of user-generated content (UGC) and manufactured questions changed across the SEO, COVID, and AI eras (\S\S\ref{sec:temporal}-\ref{sec:context}).

\begin{figure*}[t]
  \centering
  \includegraphics[width=\textwidth]{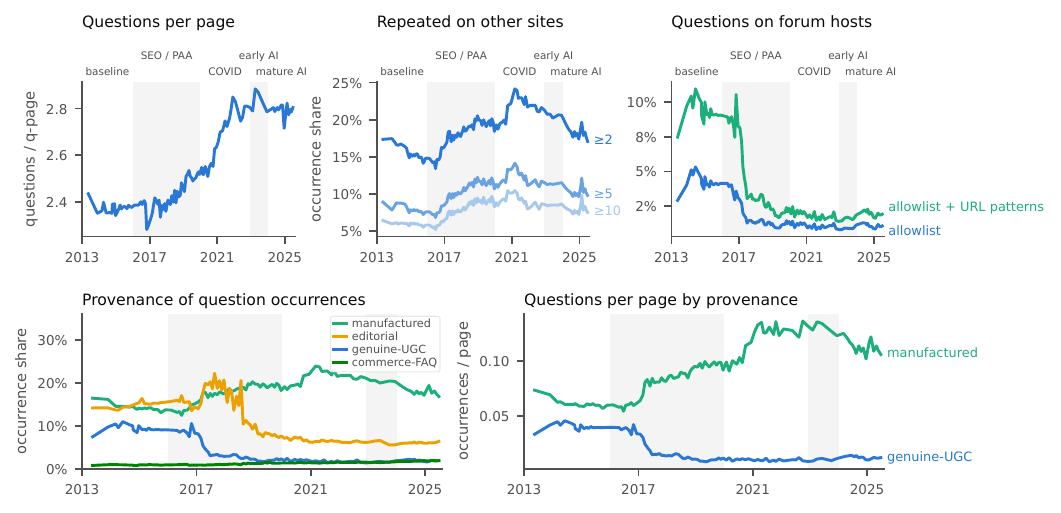}
  \caption{Twelve-year changes over 110 snapshots (\S\ref{sec:temporal}). Top: question intensity, cross-host duplication ($D{\geq}2/5/10$), and genuine-forum share (allowlist and allowlist + URL patterns). Bottom: the provenance classes (unknown omitted), first as a share of occurrences, then as questions per crawled page (does not depend on the size of unknown). Shaded bands are the five eras, named along the top row. Trends use the Hamed-Rao correction.}
  \label{fig:combined}
\end{figure*}

\section{Related work}

Web Data Commons, the closest prior work, extracts schema.org markup from one Common Crawl snapshot per year and shows FAQPage/Question adoption increasing rapidly \citep{brinkmann2023wdc,meusel2015schemaorg}, driven first by SEO then AI-search \citep{webalmanac2024structureddata}. While they count markup classes by domain, we count linguistic questions, which are the schema-less majority, and label each with provenance and form. CCQA \citep{huber2022ccqa} mines schema.org question markup, WebFAQ \citep{dinzinger2025webfaq} only FAQPage (2.0 is re-crawled \citep{dinzinger2026webfaq2}), both of which treat the result as natural asking; GooAQ \citep{khashabi2021gooaq} harvests autocomplete; PAQ \citep{lewis2021paq} is machine-generated. No prior work analyzes provenance or temporal changes at the question level.

Prior work audits web document LLM-content prevalence \citep{he2026degentweb}, the machine-generated text share over time \citep{dolezal2026aitext,sun2025aigt,thompson2024shocking}, and mass-produced SEO spam in search \citep{bevendorff2024google}, but none reaches the question layer. We also add a signal that none use: verbatim cross-host duplication (text that appears word-for-word on multiple different websites), which decreases in our study. Since others report machine text increasing, this is detector-independent evidence of a change in how questions are manufactured (\S\ref{sec:temporal}). Change in question form was last measured query-side through 2010 \citep{pang2011lostquery}, on queries typed into a search engine rather than questions written on pages, and without provenance labels; query-side work has also compared question and keyword phrasings \citep{white2015questionsqueries}. We measure it page-side, at the level of the individual occurrence, and label where each one came from. Corpus-composition auditing at scale is the closest type of work \citep{elazar2024wimbd}, as is the human audit of web-crawled corpus quality in \citet{kreutzer2022quality}; neither reaches the question layer.

\section{Data and methods}\label{sec:datamethods}

\textbf{Corpus.} We analyze questions from 110 English, quality-filtered FineWeb dumps, CC-MAIN-2013-20 through CC-MAIN-2025-26, one per Common Crawl crawl snapshot.\footnote{Hugging Face dataset \texttt{HuggingFaceFW/fineweb}, revision \texttt{9bb295ddab0e05d785b879661af7260fed5140fc} (tag \texttt{v1.4.0}, 2025-07-11) \citep{penedo2025finewebdata}. \citet{penedo2024fineweb} document the 96-dump release covering 2013 to April 2024; v1.4.0 extends it through CC-MAIN-2025-26 (June 2025) and contains the 110 dumps we process.} A \emph{question occurrence} is a sentence with a question mark of 5-100 words that starts with a question word or contains an interrogative keyword, extracted from page text and labeled with URL, host, and snapshot. We normalize before counting (NFKC, casefold, whitespace collapse, edge-punctuation strip), resulting in 13.4B occurrences with a mean length of 12.8 tokens. Our extractor was fixed and did not adapt to differences in era (see Limitations).

\textbf{Design.} We use two units of analysis throughout this work: every temporal curve and provenance share is computed over \emph{occurrences} (one question-bearing sentence on one crawled page), while the frequency analysis in \S\ref{sec:freqdemand} alone works on \emph{distinct questions} (one normalized string within a snapshot). All statistics are first computed within each snapshot then compared to others after occurrence-weighting. We never sum counts across snapshots, and subsequent percentages are relative change. Trends use a two-sided Mann-Kendall (MK) test \citep{mann1945,kendall1975}, and comparisons use five prior-frozen eras: baseline (2013-15), SEO (2016-19), COVID (2020 through CC-MAIN-2022-40), early-AI (CC-MAIN-2022-49, the ChatGPT release week, through 2023), and mature-AI (2024-25), with bootstrap CIs over per-snapshot shares (the widest era-mean half-width below is 0.8 points) and a $\pm$1-year boundary check. Consecutive snapshots are autocorrelated, so every trend below is reported under the Hamed-Rao variance correction \citep{hamed1998}, which cuts the effective series length to $n^{\ast}{\approx}9$-13; a series we describe as non-monotonic gets no whole-series trend test, since MK tests monotonic trend only. We also correct for testing many curves (App.~\ref{app:bh}). Robustness checks for crawl-size variation are in Limitations.

\textbf{Metrics.} \emph{Intensity} is occurrences per question-bearing page, which is the quantity plotted throughout. The share of pages with any question increases over the series, so this and the per-crawled-page rate diverge, and we report both. \emph{Cross-host duplication} is the share of occurrences whose normalized text appears on at least $D$ registrable domains within a snapshot, which is a conservative lower bound at $D{\geq}2$ (we do no fuzzy matching; same shape at $D{\geq}5$ and $D{\geq}10$; see Limitations on snapshot size and detection). \emph{Genuine-forum share} classifies hosts with a precision-prioritized UGC classifier, an allowlist of person-to-person hosts and URL patterns like \texttt{/forum/}, and we report separately the allowlist and allowlist + URL patterns union. \emph{Provenance} assigns each occurrence to one of five classes: genuine-UGC, manufactured-template (cross-host duplicated), commerce-FAQ, editorial, or unknown (App.~\ref{app:provenance}). \emph{Form} is a deterministic regex and part-of-speech (POS) classifier (App.~\ref{app:form}) over type, person, and scaffolding (an attached context sentence), validated against human labels (App.~\ref{app:gold}). Annotators agree on provenance at $\kappa{=}0.52$ but on authenticity (whether a question was genuinely asked) at only $\kappa{=}0.23$, which is why we focus on provenance and not authorship. We run no machine-generated-text detector since our unit is far too short for reliable detection and such detectors are fragile even on long text \citep{sadasivan2023detect}, so duplication, provenance, and form stay detector-independent. However, this leaves a large unknown residual (61\%-73\% over the series; see Limitations).

\textbf{Sampling.} The occurrence-weighted curves use the full population. Form and context being per-question properties, are estimated on a per-question sample: 20k questions per snapshot (2.2M total), drawn via a deterministic hash of the question text from files spaced across each dump. At this size a per-snapshot share carries 95\% CI half-width under 0.7 points, so the aggregated sample keeps the provenance strata large enough (28k commerce-FAQ, 92k genuine-UGC) for the form-provenance analysis in \S\ref{sec:interaction}.

\section{Form predicts provenance}\label{sec:interaction}

On a 2.2M sample with both form and provenance labels (Table~\ref{tab:interaction}), genuine questions are longer than manufactured ones (12.9 vs.\ 8.5 tokens, Cohen's $d{=}0.71$), about twice as likely to be scaffolded (7.3\% vs.\ 3.9\%), and more often to be first-person (29.7\% vs.\ 21.4\%). A logistic regression on only form features separates the two at AUC 0.725, and length is the dominant term (odds ratio 2.09 per SD). This is a population-level separation, since at the level of a single question, the model is weak, and we make no claim about individual items. It is also narrower than the pooled figure suggests. Against the commerce-FAQ class, the only class defined without duplication, AUC falls to 0.554 (App.~\ref{app:provenance}), so what form separates is genuine questions from template farms, not from commercial-FAQ writing, where only provenance gives them away. As an external check, WebFAQ, mined from \emph{FAQPage} markup, matches our manufactured class on every form measure (App.~\ref{app:form}).

We pre-specified five hypotheses before analysis (App.~\ref{app:hyp}). The one that failed was question \emph{type}, such as wh- and polar, which has almost no signal (Cramér's $V{=}0.03$); manufactured questions are stripped of length and context, and of first-person framing only weakly ($V{=}0.04$). This holds across the eras, as genuine questions stay 12.8-13.4 tokens in all, the only class whose form does not decline (App.~\ref{app:form}).

\begin{table}[h]
\centering\small
\setlength{\tabcolsep}{3.5pt}
\begin{tabular}{@{}lrrr@{}}
\toprule
 & \textbf{genuine} & \textbf{manuf.} & \textbf{effect} \\
\midrule
length (tokens) & 12.9 & 8.5 & $d{=}0.71$ \\
scaffolded      & 7.3\% & 3.9\% & $\phi{=}0.07$ \\
first-person    & 29.7\% & 21.4\% & $\phi{=}0.08$ \\
\bottomrule
\end{tabular}
\caption{Form by provenance; 2.2M sample ($n{=}511$k: 92k genuine, 420k manufactured). \emph{Scaffolded}: the question has an attached context sentence. Manufactured $=$ template $\cup$ commerce-FAQ. Effects are genuine vs.\ manufactured, where length does most of the work; App.~\ref{app:hyp} reports Cramér's $V$ for person and type over all five classes.}
\label{tab:interaction}
\end{table}

\section{Frequency is not a reliable proxy for demand}\label{sec:freqdemand}

A natural way to read demand from the web is to count, but we find this to be flawed. All counts here are within-snapshot occurrence counts of normalized text, on three snapshots: 2015, 2020, 2024 (one in the middle of each era).

We rank each snapshot's distinct questions by count. The top-1{,}000 are 71.7\%, 86.8\%, and 93.2\% manufactured-template in 2015/2020/2024, against 3-4\% of distinct questions corpus-wide. The genuine-UGC share of that is 6.3\% in 2015 and 0.3\% in both 2020 and 2024. The most-verbatim-repeated strings are boilerplate: ``what are you waiting for'' (72{,}331 occurrences in 2020), ``how can we help you''. Furthermore, genuine demand never concentrates in the distribution, with 97\% of distinct questions occurring once or twice and the top-1{,}000 are under 3\% of all occurrences. Therefore, occurrence frequency measures publishing intensity. We make no claim in this work about what real demand is.

\begin{figure}[t]
  \centering
  \includegraphics[width=\columnwidth]{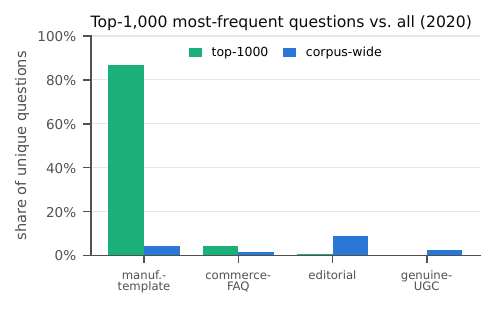}
  \caption{Provenance of the top-1{,}000 most-frequent questions vs.\ all distinct questions (2020; \emph{unknown} omitted). Of the top-1{,}000, 86.8\% are cross-host duplicated strings, against 4.1\% of distinct questions corpus-wide.}
  \label{fig:freqdemand}
\end{figure}

\section{Who's asking, over time}\label{sec:temporal}

Figure~\ref{fig:combined} summarizes the twelve-year changes.

\textbf{Intensity} rises from an era mean of 2.37 in 2013-15 to a 2.83 peak in 2022-23, settling at 2.79 by 2024-25 ($+17.7\%$ over twelve years), concentrated in the SEO and COVID eras when FAQ adoption accelerated as previously mentioned. The share of pages with any question (18.4\% to 22.5\%, 2013-2025) moves independently of FineWeb's per-snapshot retention (Spearman $\rho=-0.08$). Since more pages have questions and each has more of them, questions per crawled page increase $+46.6\%$.

\textbf{Cross-host duplication} is non-monotonic: 15.7\% of occurrences at baseline, up to 21.6\% at a peak in CC-MAIN-2021-10, then down to 18.3\% through the AI eras (same shape at $D{\geq}5/10$). We therefore do not report a whole-series trend for this curve. Duplication is also a detection event, and snapshot size varies 3.6$\times$ over the series, so we hold size fixed: each mature-AI snapshot duplicates less than the 2020-21 snapshot nearest it in size, 16 times out of 16 (mean $-3.5$ points). The peak precedes the ChatGPT release by around twenty months. If questions were generated by simply copying templates, duplication would keep increasing. Instead, AI-era text appears novel, so verbatim duplication would undercount manufactured questions after 2022. Detector-based studies find machine content rising over the same window \citep{he2026degentweb,dolezal2026aitext}.

\textbf{Genuine-forum share} collapses on both definitions: the allowlist from 4.2\% to 1.1\% ($-74.3\%$) and the broader allowlist + URL pattern union from 9.4\% to 2.0\% ($-79.0\%$). On a fixed panel of frequent hosts, the declines shrink to $-42\%$/$-56\%$ (see Limitations). Its recall on human-labelled UGC is lowest in the baseline era (App.~\ref{app:gold}), which would understate the decline rather than inflate it. Most of the fall precedes the 2023-24 crawler-blocking wave \citep{longpre2024consent} and holds on pre-2023 snapshots, so it is not confounded by forums leaving the crawlable web (see Limitations).

\textbf{Provenance.} From 2013 to 2025, genuine-UGC falls from 9.3\% to 1.9\% ($-79.2\%$), manufactured-template rises $+25.9\%$ (same peak as the duplication curve), and commerce-FAQ doubles from a small base ($+102\%$). These are shares of all occurrences, so a growing unknown class would shrink them on its own. We therefore also measure each class per page: genuine-UGC falls from 0.222 to 0.054 occurrences per question-bearing page ($-75.5\%$), and from 0.040 to 0.012 per crawled page ($-69.5\%$), while manufactured-template rises $+84.6\%$ and commerce-FAQ $+196\%$ per crawled page. Unknown accounts for at most ten of the seventy-nine points, so renormalization is not the cause of the decline. Crawl size is a confound, and adjusting for it, the manufactured-template \emph{share} rise does not survive ($-2.0\%$, $p{=}0.49$) while the per-page rate does ($+13.3\%$, $p{=}0.006$). We therefore claim the growth of the manufactured layer as a rate and not as a share (see Limitations). This is corroborated by a simultaneous rise in FAQ-path URLs (\texttt{/faq}, \texttt{questions-and-answers}) from 0.82\% to 1.42\% ($z{=}{+}6.6$), steepest from 2017-19, the same window of rapid \emph{FAQPage} adoption. However, this counters the idea of ``question farms'', as the top-100 hosts' share \emph{falls} from 11.7\% to 2.8\%.

\textbf{Form}. Mean length falls from 13.41 to 11.85 tokens ($-11.6\%$) and the scaffolded share from 9.95\% to 6.12\% ($-38.5\%$). What diminishes is the human around the question, such as a background sentence or first-person framing, which is the same pattern we see in \S\ref{sec:interaction}. The remaining genuine questions are verbose while the growth is in stripped templated strings, matching studies of online Q\&A community volume collapsing \citep{delriochanona2024llmqa,burtch2024genai}; on Stack Overflow, the questions that still get asked are longer and harder to answer \citep{helic2026stackoverflow}.

\section{Context-independence}\label{sec:context}

Most of the questions we observe are not self-contained; under a deterministic POS-based rubric (self-contained / conversational-fragment / context-dependent), about half of our sample are self-contained, 38\% are conversational fragments, and 12\% carry unresolved references. A human check agrees with the rubric on 63\% of items, and the needs-context flag has precision 0.75 (App.~\ref{app:context}), so we read this section as exploratory and its rates as directional. Over time, the context-dependent share decreases from 14.0\% to 8.6\%, which holds under the Hamed-Rao correction; the matching rise in the self-contained share does not ($p{=}0.20$), so we do not make a claim about it.

Genuine-UGC is the most context-dependent class (19\%); we interpret this as real questions being written for people to read, so they depend on surrounding context. Commerce-FAQ is the most self-contained (59\%), because an FAQ page's purpose is to present and answer a standalone question. Filtering a web corpus for clean, standalone questions therefore preferentially keeps the manufactured questions.

\section{Discussion}

A growing share of the web's readable questions is there to be matched by search engines and now generative answer engines. Systems that interpret web question frequency as demand at face value are sampling the manufactured layer of short, templated, impersonal, standalone strings. When retrieval for such systems draws on this layer, quality has been shown to degrade in simulation \citep{yu2026retrievalcollapse}.

The manufactured layer is full of fluent and well-formed text on the surface. Telling the genuine from manufactured requires looking at provenance, i.e. where a question occurs, how common it is, what form it takes, and its context. Frequency does not measure demand on its own (\S\ref{sec:freqdemand}).

Today's manufactured web was built by the SEO economy well before 2022. After that we see the signal weaken: verbatim cross-host duplication, which caught manufactured questions for a decade, catches less of them in the AI eras, and its peak precedes the ChatGPT release by around twenty months. We run no detector, so we cannot attribute that to model-written text. Either way, duplication no longer finds the manufactured layer, and future measurement will need a different signal.

\section*{Limitations}

\textbf{Corpus scope.} Our curves are measured on the FineWeb-filtered English web. A re-extraction from raw Common Crawl WET text (three dumps spanning the range: CC-MAIN-2015-40, -2020-34, and -2024-51) reproduces the twelve-year directions web-wide (intensity up, forum share down from a 12.7\% raw base in 2015), but the raw and filtered curves diverge after 2020. We therefore make the twelve-year claims web-wide and confine the post-2020 contrasts (early-AI vs.\ mature-AI) to the FineWeb slice. Our analysis is restricted to English; trends may differ across languages.

\textbf{Filter-induced exclusion.} Our design cannot separate a decline in genuine UGC on the filtered web from a quality filter that increasingly excludes it. The raw 2015 forum base is 12.7\% against 9.4\% filtered, and quality filters are documented to select against informal registers and community-authored text \citep{dodge2021c4,kreutzer2022quality}. Part of what we measure as genuine UGC leaving the corpus may be the filter declining to keep it. This matters beyond our estimates, because the excluded material skews toward the same non-standard registers the field has been trying to stop under-representing. The raw re-extraction that bounds this rests on three dumps, which is enough to check direction and not enough to correct magnitude.

\textbf{Measurement.} Our genuine-UGC detection values precision over recall, so the genuine shares are lower bounds and we claim trends only and genuine text on unrecognized hosts is left to the \emph{unknown} class. Shares are computed over all occurrences, so a growing unknown residual shrinks every classified share \citep{aitchison1982}, and part of a measured decline could be questions moving into unknown. Measured per page instead, genuine-UGC still falls $-75.5\%$ per question-bearing page and $-69.5\%$ per crawled page against the $-79.2\%$ share, so the residual accounts for at most ten of those points. Of 49 human-labelled unknown items only 4 were genuinely UGC; reassigning that 8\% of the residual to genuine gives a $-45.2\%$ decline that still holds under the correction ($p{=}0.01$), and the decline would only disappear if a quarter of the residual were genuine, which is three times what the sample supports. Unknown's form also de-scaffolds over time like the manufactured and editorial classes, while genuine UGC stays flat. The unknown class remains the weakest part of the scheme: it holds 61\%-73\% of occurrences at precision 0.43, so we treat every classified share as a lower bound rather than a partition of the web's questions. Because we run no short-text machine-generated-text detector (\S\ref{sec:datamethods}), the manufactured class inherits verbatim duplication's post-2022 undercounting, and we claim only that our AI-era findings are consistent with a shift to AI-generated text and not that any question was model-written. The fixed extractor cannot adapt to era-specific styles, so recall drift could in principle mimic a trend; but the forms it under-captures (elliptical and noisy informal questions) skew genuine, so its misses leave our genuine-share estimates conservative. Form labels are heuristic and evaluated against human labels (App.~\ref{app:gold}).

\textbf{Crawl artifacts.} The host-level measurements could move with what Common Crawl fetches rather than with the web. Which hosts it fetches shifts over the years \citep{thompson2024longitudinal}; on a fixed panel of the 79{,}826 hosts present in $\geq$90\% of snapshots the forum declines are still the same direction but shrink ($-42\%$/$-56\%$ vs.\ $-74\%$/$-79\%$ crawl-wide) and top-100 concentration falls $-57\%$, so the direction is a genuine web trend. The decline is smaller on the panel, and that gap, about 40\% of the full-crawl figure, comes from changes in which hosts Common Crawl fetches. Major UGC platforms have also restricted open crawlers since 2023 \citep{longpre2024consent}, so late-era genuine levels reflect only the crawlable web, but recomputing on pre-2023 snapshots shows withdrawal does not drive the decline and is if anything steeper. Consecutive snapshots are autocorrelated since under the Hamed-Rao correction the effective series length falls to $n^{\ast}{\approx}9$-13, yet intensity, both forum signals, and shallow duplication ($D{\geq}2$) stay significant ($p<0.02$), while $D{\geq}5$ is marginal and $D{\geq}10$ non-significant. Thus, we claim duplication only at the shallow threshold.

\textbf{Crawl size.} Snapshot size varies 3.6$\times$ over the series (57.6M to 206M occurrences), and duplication is a detection event, so larger snapshots find more of it: the first differences of the duplication and size series correlate at $\rho{=}{+}0.66$, against $|\rho|{\leq}0.33$ for every other curve. Regressing each curve on log size and re-levelling at the mean, the twelve-year contrasts shrink and two reverse. The apparent net increase in duplication does not survive (from $+16.6\%$ to $-4.7\%$, $p{=}0.81$), and neither does the manufactured-template share's (from $+25.9\%$ to $-2.0\%$, $p{=}0.49$); we make neither claim. What survives is intensity ($+6.9\%$), commerce-FAQ ($+46.4\%$), the manufactured-template rate per crawled page ($+13.3\%$, $p{=}0.006$), top-100 concentration ($-39\%$), the form drift, and the genuine decline at $-45.0\%$, which is where the fixed host panel also lands ($-42\%$/$-56\%$). Two independent controls converging on roughly $-45\%$ is our best estimate of the genuine decline; the crawl-wide $-79\%$ is an upper bound on it. The size-adjusted genuine series does not clear the Hamed-Rao threshold on its own ($p{=}0.13$). With $n^{\ast}{\approx}9$ and two corrections applied in sequence, we are at the limit of what 110 snapshots can support.

\section*{Ethics and artifacts}

The corpus is derived from Common Crawl via FineWeb (ODC-By). We analyze public web text in aggregate and release no personal data. The validation sample is constructed to be PII-clean, with candidates containing a person name, email, or phone number being dropped at selection and replaced from the same stratum, and was screened for sensitive topics by a keyword filter plus a manual read of every item. Host and URL are withheld from every released artifact, so no released string is linkable to a page or author. At \url{https://github.com/bodhiumlabs/tell-whos-asking} we release the validation sample and its guidelines, the UGC host classifier (App.~\ref{app:ugcclf}), the commerce and editorial host lists and hostname patterns, the interpretable form and context classifiers, the LLM-judge script (prompt, model id, and decoding settings), per-snapshot denominators, the fixed host panel, per-figure curve data, and the deterministic rule that selects the 2.2M sample (the fingerprint-modulus predicate over question text, the modulus, and the per-dump file spacing), which rebuilds the same sample from public FineWeb without releasing document identifiers. The per-occurrence text stays unreleased. The validation sample contains no content from hosts that signalled crawler opt-out at the time of release \citep{longpre2024consent}.

\textbf{Funding and competing interests.} All authors are affiliated with Bodhium Labs, which develops commercial products in answer- and generative-engine optimization. The work was funded by Bodhium Labs, which also provided the compute.

\textbf{Dual use.} A released classifier that separates genuine from manufactured questions can also be used to evade it. The changes \S\ref{sec:interaction} implies are cheap with an LLM: lengthen the question, attach a context sentence, write in the first person. If content farms make them, this measurement stops working, so we expect it to date quickly. We release anyway because the same detail lets dataset builders audit what they have collected. Read the other way, our decline covers the crawlable, quality-filtered layer only, and taking it as a verdict on community-generated content would compound the exclusion described above.

\section*{Acknowledgments}

We thank Ruben Ramirez Salas and Alex Miller for their comments on earlier drafts, and the four anonymous reviewers for the depth of their reviews.

\bibliography{custom}

\appendix

\section{Question datasets}\label{app:landscape}

Table~\ref{tab:landscape} places this work among representative question datasets by provenance family. Each family carries a distinct limitation, i.e. why its questions are not a clean record of human demand:

\begin{description}\itemsep2pt
\item[Query logs] record real asking, but as \emph{mediated clicks}; the true frequencies that would make them a distribution were withdrawn (AOL) or stripped (ORCAS).
\item[Community-QA] is human-authored but gated or deduplicated, and its vote signal ranks only within a single forum.
\item[Markup mining] takes site-published \texttt{schema.org} QA annotation: WebFAQ restricts to \texttt{FAQPage}, while CCQA's \texttt{Question} mixes those with community asking (Stack Exchange is its largest source), unlabelled.
\item[Autocomplete/PAA] harvests search-\emph{suggested} completions, not questions anyone wrote directly.
\item[Synthetic] generation has no human denotation at all and is effectively unbounded.
\item[Chat logs] capture people \emph{prompting an assistant}, not asking on the open web.
\item[Full-text mining] (this work) reads question sentences off pages at web scale, so it reaches the whole distribution, but its composition must be measured.
\end{description}

\begin{table}[h]
\centering\footnotesize
\setlength{\tabcolsep}{3.5pt}
\resizebox{\columnwidth}{!}{%
\begin{tabular}{@{}lllrll@{}}
\toprule
\textbf{Dataset} & \textbf{Yr} & \textbf{Family} & \textbf{Size} & \textbf{Freq} & \textbf{Access} \\
\midrule
MS MARCO WS \citep{chen2024msmarcoweb} & '24 & query log & $10^{7}$ & dist. & open \\
ORCAS \citep{craswell2020orcas}        & '20 & query log & $10^{7}$ & no  & open \\
AOL log \citep{pass2006aol}            & '06 & query log & $10^{7}$ & yes & withdrawn \\
Yahoo!\ L6 \citep{yahoowebscopeL6}     & '07 & CQA        & $10^{6}$ & no  & gated \\
Quora \citep{iyer2017quora}            & '17 & CQA        & $10^{5}$ & no  & open \\
GooAQ \citep{khashabi2021gooaq}        & '21 & autocompl. & $10^{6}$ & no  & open \\
CCQA \citep{huber2022ccqa}             & '22 & QA markup  & $10^{8}$ & no  & code \\
WebFAQ \citep{dinzinger2025webfaq}     & '25 & FAQ markup & $10^{8}$ & no  & open \\
PAQ \citep{lewis2021paq}               & '21 & synthetic  & $10^{8}$ & no  & open \\
Persona Hub \citep{ge2024personahub}   & '24 & synthetic  & $10^{9}$ & no  & partial \\
WildChat \citep{zhao2024wildchat}      & '24 & chat log   & $10^{6}$ & no  & terms \\
LMSYS-1M \citep{zheng2024lmsyschat}    & '24 & chat log   & $10^{6}$ & no  & terms \\
\midrule
\textbf{This work} & '25 & full-text & $10^{10}$ & counts$^{\ast}$ & aggr.\ only \\
\bottomrule
\end{tabular}}
\caption{Question datasets by provenance family (compressed; representative members). \textbf{Size}: order of magnitude of questions/queries. \textbf{Freq}: whether the dataset carries a frequency/popularity signal; \emph{yes} = per-query counts released, \emph{dist.} = the sample preserves the query distribution but not per-query counts, \emph{no} = a deduplicated list. $^{\ast}$We report within-snapshot occurrence counts, which is not demand. \emph{aggr.\ only}: we release per-figure curve data and the validation sample and not the per-occurrence text.}
\label{tab:landscape}
\end{table}

\section{Provenance classes}\label{app:provenance}

Each question occurrence is labeled by an \emph{ordered cascade} over its host and URL where the first matching rule wins. In order, each test applying only to occurrences the previous ones did not take:

\begin{enumerate}\itemsep0pt\parsep0pt
\item UGC host set, or person-to-person URL pattern? $\to$ \texttt{genuine-UGC}
\item normalized text on $\ge2$ registrable domains? $\to$ \texttt{manufactured-template}
\item host is commerce or support/FAQ? $\to$ \texttt{commerce-faq}
\item host is news, blog or media? $\to$ \texttt{editorial}
\item otherwise $\to$ \texttt{unknown}
\end{enumerate}

\noindent The classes are therefore mutually exclusive by construction, and \texttt{unknown} is reached only by failing all four tests. Duplication is computed on the question text normalized by NFKC, casefold, whitespace collapse, and edge-punctuation stripping (as in \S\ref{sec:datamethods}). The rules in full:

\begin{description}
\item[\texttt{genuine-UGC}] Host is in a user-generated-content set: the $\sim$1M most frequent hosts (by occurrence) run through a forum/UGC host classifier, \emph{or} the URL matches a person-to-person pattern (\texttt{/forum}, \texttt{/thread(s)}, \texttt{/topic}, \texttt{/comments}, \texttt{/discussion}, \texttt{/board}, \texttt{viewtopic.php}, \texttt{showthread.php}, or a \texttt{forum.}/\texttt{ask.}/\texttt{community.}/\texttt{discuss.} subdomain). We tested this first, and it works even for duplicated text. This is a lower bound, since only recognized UGC hosts/patterns are caught, so genuine questions on unrecognized hosts fall through.

\item[\texttt{manufactured-template}] Among non-UGC pages, the normalized question occurs on $\ge2$ distinct registrable domains: cross-host duplication (canonical FAQ/People-Also-Ask/boilerplate republished across sites). This is defined by \emph{duplication} and not by wording or length.

\item[\texttt{commerce-faq}] Non-UGC, non-duplicated; host is a commerce or support/FAQ site, identified first from a curated list of hosts labelled by category, and for hosts not on the list from hostname patterns (\texttt{shop|\allowbreak store|\allowbreak cart|\allowbreak buy|\allowbreak deals|\allowbreak coupon|\allowbreak pricing|\allowbreak ecommerce}; \texttt{support|\allowbreak help|\allowbreak faq|\allowbreak docs|\allowbreak kb|\allowbreak answers|\allowbreak knowledgebase}).

\item[\texttt{editorial}] Host is a news/blog/magazine/media outlet (\texttt{news|\allowbreak blog|\allowbreak magazine|\allowbreak times|\allowbreak post|\allowbreak herald|\allowbreak journal|\allowbreak media|\allowbreak wire}).

\item[\texttt{unknown}] No rule matched: not recognized UGC, not duplicated, and the host in no category. This is the residual, unclassifiable layer, whose share increases from 61\% to 73\% over the series; whether that shrinking base re-labels the classified trends is in Limitations.
\end{description}

These hostname patterns match substrings, so some of them collide: \texttt{post|times|wire} also catches \emph{poster} and \emph{wireless}, and \texttt{answers|docs|kb} also catches community Q\&A and technical documentation. We re-ran the host-type labeling with those alternatives dropped, on one snapshot per year. The editorial trend is unaffected ($-63.1\%$ against $-64.6\%$ pruned), while the FAQ trend roughly doubles ($+54.9\%$ against $+115.0\%$), so dropping the collisions strengthens the FAQ trend instead of weakening it. These are host-type shares, not the cascade classes, because duplication is tested first; they show how much the collisions could matter, not what \S\ref{sec:temporal} reports.

Provenance is decided from host and cross-host duplication, and never from a question's length, scaffolding, or type. Therefore, the form-to-provenance detector (\S\ref{sec:interaction}) is not circular. However, one residual mechanism remains: manufactured-template is defined by cross-host duplication, and short strings duplicate more readily, so ``length predicts manufactured'' partly restates ``short strings recur.'' We examine this by refitting on the \texttt{commerce-faq} subset, defined by host category rather than duplication. Here form barely separates the classes (AUC 0.554, against 0.725 for the main fit), so dropping the duplication-defined template class takes most of the length signal with it. So the form-to-provenance detector separates genuine questions from template farms, not from commercial-FAQ language, and length carries what separation remains. We report this as the detector's boundary; the provenance and temporal results (\S\S\ref{sec:freqdemand}-\ref{sec:temporal}) do not depend on it.

\section{Form over time}\label{app:form}

The form classifier is deterministic regex and POS rules (one-shot, no training data) labelling three independent dimensions of a raw question string:

\begin{description}
\item[\texttt{type}] a priority cascade, first match wins: \texttt{tag} (declarative with a $\le$3-token tag after a comma, ``\dots, right?'') $>$ \texttt{imperative-interrogative} (clause-leading base-form verb) $>$ \texttt{wh} (a wh-tagged token) $>$ \texttt{polar} (leading auxiliary or modal) $>$ \texttt{elliptical} (no finite verb but a contentful noun/verb phrase) $>$ \texttt{other}.
\item[\texttt{person}] \texttt{first} if a first-person pronoun (\texttt{i/we/my/our\dots}) appears, else \texttt{second} (\texttt{you/your}), else \texttt{impersonal}.
\item[\texttt{scaffolding}] present if the question spans more than one sentence, opens with a subordinator (\texttt{because}, \texttt{if}, \texttt{when}, \dots), or carries a politeness token (\texttt{please}, \texttt{thanks}, \dots); length is the content-token count.
\end{description}

\begin{figure}[h]
  \centering
  \includegraphics[width=\columnwidth]{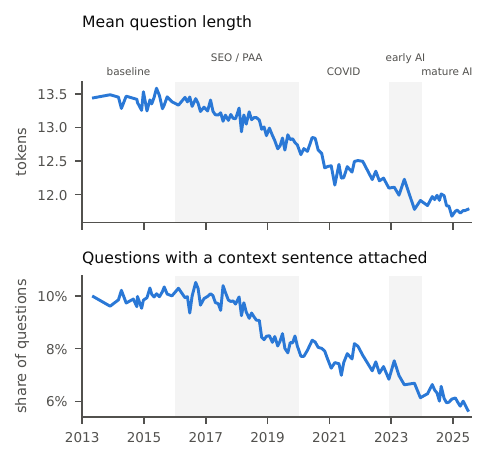}
  \caption{Form over time: mean length falls from 13.41 to 11.85 tokens ($-11.6\%$) and the scaffolded share falls from 9.95\% to 6.12\% ($-38.5\%$); both monotonic, MK $p{<}10^{-4}$. (Hamed-Rao $p{<}0.002$). Percentages are computed on unrounded era means.}
  \label{fig:form}
\end{figure}

Aggregated over the 2.2M sample, questions are 73.0\% \texttt{wh} and 24.3\% \texttt{polar} (\texttt{tag} 0.6\%, \texttt{imperative} 1.0\%, \texttt{elliptical} 0.2\%, \texttt{other} 0.9\%); 43.9\% impersonal, 32.6\% second-person, 23.6\% first-person; 8.5\% scaffolded; mean length 12.8 tokens. Figure~\ref{fig:form} plots the two monotonic aggregate trends behind \S\ref{sec:temporal}; the type distribution stays flat (\texttt{elliptical} a 0.2\% floor).

Separated by provenance, every class shortens and de-scaffolds monotonically across the five eras \emph{except} genuine-UGC, which stays flat (from 12.8 to 13.4 tokens; scaffolding from 7.0\% to 8.5\%, non-monotonic) while manufactured-template (from 8.6 to 7.8), commerce-FAQ (from 13.3 to 11.5), editorial (from 14.9 to 13.6), and unknown (from 14.2 to 12.8) all fall. So the corpus-level form decline is within-class degradation plus composition shift and not composition alone. Genuine-UGC's sampled base diminishes over the same window (from ${\sim}38$k to ${\sim}6$k occurrences), consistent with its decreasing share, so we read its trajectory as flat-within-noise rather than increasing.

Length dominates the \S\ref{sec:interaction} model, so we ask what remains once it is removed. On 60k questions per class (all 110 snapshots, spaCy tagger) we compute POS rates and lexical diversity, reweighting every class to the genuine-UGC length distribution over seven length bins so each comparison is made at equal length. Grammatically, the strata are close (manufactured vs.\ genuine at equal length: adjectives $-4\%$, nouns $-2\%$, verbs equal, content-word density $-3\%$); they part on reference and repetition: proper nouns $+35\%$ (brand and product names; commerce-FAQ is the extreme at .084 vs.\ genuine .041), pronouns $+14\%$ (second-person calls to action), and type-token ratio (TTR) over a fixed 200k-token budget $11\%$ lower. The template class's low TTR (.069 vs.\ genuine .090) partly restates its duplication-based definition. Commerce-FAQ, which is not as defined, is at .094, so we take the conservative reading. Manufactured questions are built much like genuine ones but name different things and repeat themselves more. Deeper syntactic and discourse analysis would require a parser pass that we did not run.

An external cross-check corroborates the manufactured form signature. WebFAQ \citep{dinzinger2025webfaq,dinzinger2026webfaq2}, mined from schema.org \emph{FAQPage} markup, matches our manufactured class on every measure (8.96 tokens vs.\ our manufactured 8.5 and genuine 12.9; 4.0\% scaffolded, 74.7\% impersonal). This is evidence that markup mining harvests manufactured questions and treats them as natural asking. This is form-only, so WebFAQ's small-business host tail does not overlap our site-type labeler, so we do not compare provenance.

\section{Human validation}\label{app:gold}

A human validation sample stratified by provenance class and era is labelled by three annotators: a shared 50-item calibration block for a three-way inter-annotator overlap, plus 100 disjoint items split across the three, 150 distinct items in all ($\sim$83 per annotator). Each item is shown with the page's \emph{as-crawled} text, so pages that have since died or been repurposed are judged from their original content. Annotators judge three cues per question \emph{with its host in view}, and authenticity (genuine / manufactured / unclear) is computed from them:

\begin{description}
\item[community] the host is a person-to-person venue (forum, board, comment thread, Q\&A), judged from the host rather than whether the wording sounds human.
\item[templated] the exact string could recur across many sites (canonical FAQ, SEO or marketing boilerplate), which is a template signal, independent of length or person.
\item[personal] first-person, or about the asker's own situation.
\end{description}

A blind LLM rater (\texttt{gemini-3.5-flash}, run 2026-08) applies the same cues to each question's text, host, URL, and as-crawled page text, but never sees the human labels, never sets one, and only cross-checks them. The annotators are the authors, so this blind rater is the independent check on the labels, which is why we report inter-annotator $\kappa$ and human-LLM agreement rather than treat the human labels as ground truth. Because provenance is decided from host and cross-host duplication, not from a question's form (App.~\ref{app:provenance}), the form-to-provenance detector is not circular; the human sample checks its target rather than restating it.

The human sample validates the detector's target: classifier provenance matches adjudicated human provenance at overall precision $0.73$ (Wilson 95\% CI $[0.66,0.80]$), with genuine-UGC precision $1.00$ ($n{=}29$); $[0.88,1.00]$, so the lower-bound reading of the genuine share tolerates a true precision near 0.88, editorial $0.93$, manufactured-template $0.80$, and commerce-FAQ $0.74$; the unknown residual is lower ($0.43$, $n{=}49$), as expected of a catch-all. Provenance inter-annotator $\kappa$ is $0.52$, which is moderate agreement \citep{landis1977}. Per-class figures at $n{=}150$ are indicative since we read the intervals.

Era drift is the main issue when it comes to a twelve-year trend, so we also separate precision out by era (the sample is stratified on era, $n{=}25$-35 each). Excluding the unknowns, per-era precision is $0.81$/$0.95$/$0.90$/$0.83$/$0.93$ from baseline to mature-AI, and genuine-UGC precision is $1.00$ in every era (8/8, 6/6, 6/6, 5/5, 4/4). On the recall side, of the 37 items human annotators called genuine UGC the classifier caught 29 ($0.78$), lowest in the baseline era (8/11 vs.\ 4/5 in mature-AI), which is the direction that leaves the measured decline conservative, though at 5-11 items per era this is weak evidence and we treat it as such.

Authenticity itself (genuine / manufactured / unclear) is a harder, more subjective judgment: three-way $\kappa$ is only $0.23$. The disagreement is not random since agreement rises with how decisive the cues are: on the 50 shared items, annotators agree unanimously on authenticity for 60\% of items where all three cues match ($n{=}10$), 46\% ($n{=}24$), 21\% ($n{=}14$), and 0\% where none match ($n{=}2$). It also varies by host type, from 79\% on genuine-UGC items to 25\% on editorial, 14\% on commerce-FAQ and 0\% on the unknown class. Annotators split on the items whose host gives no clear signal, so the difficulty is in the construct and not the annotation. These counts are small, and even where all three cues match two in five items still split. Manufactured questions now read like genuine ones, which is the paper's central finding. What stays legible, and what the detector targets, is the host-level provenance these questions carry. We therefore base validation on provenance (which annotators \emph{do} agree on) and report no form-to-authenticity refit. Form fields (type, person, scaffolding) were left almost entirely uncorrected against the classifier ($>{}0.98$ agreement); human-LLM agreement is likewise modest ($\kappa{=}0.38$, above the human-human level, and well below the agreement strong judges reach on easier constructs \citep{zheng2023judge}), an independent sign that the construct is the hard part. (The context rubric is validated by a single-annotator spot-check on a separate era-balanced sample, App.~\ref{app:context}.)

\section{Pre-specified hypotheses}\label{app:hyp}

The five directional hypotheses for the form-by-provenance test, fixed before the joint distribution was computed:

\begin{description}
\item[H1 (person)] genuine-UGC is more first-person than manufactured. \emph{Weak}: 29.7\% vs.\ 21.4\% (Cramér's $V{=}0.04$).
\item[H2 (scaffolding)] genuine-UGC is more scaffolded. \emph{Held}: 7.3\% vs.\ 3.9\% ($\phi{=}0.07$).
\item[H3 (length)] genuine-UGC is longer. \emph{Held}: 12.9 vs.\ 8.5 tokens ($d{=}0.71$).
\item[H4 (type)] manufactured skews to templated/elliptical/impersonal forms, genuine to fuller wh-questions. \emph{Failed}: question type carries almost no signal ($V{=}0.03$).
\item[H5 (predictive)] form features predict the genuine-vs-manufactured binary above chance. \emph{Held}: AUC 0.725 (balanced class weights; 18\%/82\% class split).
\end{description}

\section{Multiple comparisons}\label{app:bh}

We report trends for many curves, so we apply Benjamini-Hochberg \citep{benjamini1995} at $q{=}0.05$ across all 16 of them, on top of the Hamed-Rao correction each $p$-value already has. While fifteen survive, the one that does not is duplication at $D{\geq}10$ ($p{=}0.061$, $q{=}0.061$), which we already decline to claim; $D{\geq}5$ sits on the boundary ($q{=}0.050$). Ordered by corrected $p$, the classified provenance trends and the per-page rates clear the threshold comfortably (commerce-FAQ, the unknown class, manufactured-template per page, intensity and the two forum signals all at $q{\leq}0.004$), and the weakest surviving tests are genuine-UGC per crawled page and editorial per crawled page ($q{=}0.021$). Multiplicity therefore changes no claim in this paper. We note that this is a correction for testing many curves, and is independent of the crawl-size confound, which is what actually moves the duplication and manufactured-template share results (Limitations).

\section{The UGC host classifier}\label{app:ugcclf}

The genuine-UGC classifier is not a trained model: it is a deterministic function of the host string alone, with no training data and no threshold to tune, applied to the $\sim$1M most frequent hosts by occurrence; the long tail's UGC is caught by the URL patterns in the cascade (App.~\ref{app:provenance}).

A host is granted \texttt{genuine-UGC} only by (a) membership in a curated set of known person-to-person platforms, or (b) one of a small number of structural signals: a \texttt{forum}/\texttt{community}/\texttt{boards}-style subdomain, \texttt{forum} inside the registrable domain, or a forum-shaped URL path. Two rules make it precision-first. A bare \texttt{answers} or \texttt{ask} substring never grants genuine-UGC, since those are the strings SEO Q\&A mills hide behind; such hosts route to a separate content-farm category. Anything else returns \texttt{unknown} rather than a speculative label, which is why the residual is large.

Because the classifier is deterministic over host strings, it has no era-dependent behaviour by construction. Its only channel for era-varying recall is drift in how hosts are named over twelve years. We cannot rule that out, and we do not have a host-level recall measurement stratified by era. What we have is the question-level check in App.~\ref{app:gold}: recall on human-labelled genuine UGC is 0.78 overall and lowest in the baseline era (8/11, against 4/5 in mature-AI). At 5-11 items per era this is weak evidence, though the bias would understate the decline rather than inflate it. We release the classifier source, so the rules can be checked directly.

\section{Context-independence}\label{app:context}

Each question gets one of three labels by a deterministic POS-based rubric (spaCy tagger only; no parser or lemmatiser, so rules read token text, POS, and position), first match wins:

\begin{description}
\item[\texttt{conversational-fragment}] second-person deixis (\texttt{you}/\texttt{your}), or a discourse-marker opener (\texttt{so}, \texttt{and}, \texttt{but}, \texttt{ok}, \texttt{well}) plus a comma, like someone mid-exchange.
\item[\texttt{context-dependent}] an unresolved reference outside the question: a dangling demonstrative (\texttt{this}/\texttt{that}/\texttt{these}/\texttt{those} as determiner or pronoun with no following head noun), an anaphoric \texttt{it} with no prior noun antecedent, or a definite reference to an unstated artifact (\texttt{the above}, \texttt{the following}, \texttt{either}).
\item[\texttt{self-contained}] none of the above.
\end{description}

\begin{figure}[h]
  \centering
  \includegraphics[width=\columnwidth]{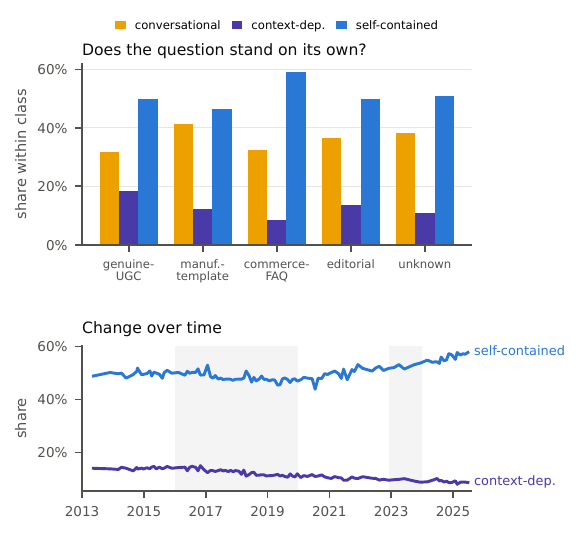}
  \caption{Context-independence of web questions. Top: label composition within each provenance class. Bottom: the context-dependent share falls over twelve years (significant under Hamed-Rao) while the self-contained rise is not ($p{=}0.20$)}
  \label{fig:context}
\end{figure}

Being POS-only, the rubric errs in both directions: it over-counts (expletive ``it is true that\dots'' reads as anaphoric) and under-counts open-class definite references (``the video'', ``both documents'') that fall outside its fixed \texttt{the above}/\texttt{the following}/\texttt{either} list. A blind, era-balanced single-annotator spot-check ($n{=}60$) performed by the author agrees with the rubric on the self-contained vs.\ needs-context split for 63\% of items (the needs-context flag is 0.75-precise), so we read the per-class rates below as directional. Figure~\ref{fig:context} gives the full within-class cross-tab and the two diachronic trends behind \S\ref{sec:context}: genuine-UGC is the most context-dependent class (18.6\%), commerce-FAQ the most self-contained (59.1\%), and manufactured-template the most conversational (41.5\%, second-person call-to-action boilerplate).

\end{document}